\documentclass[runningheads]{llncs}
\usepackage{hyperref}
\usepackage{times}
\usepackage{graphicx}
\usepackage{amsmath}
\usepackage{amssymb}
\usepackage{mystyle}
\usepackage[]{algorithm2e}
\usepackage{color}
\usepackage[width=122mm,left=12mm,paperwidth=146mm,height=193mm,top=12mm,paperheight=217mm]{geometry}
\DeclareGraphicsExtensions{.pdf,.png,.jpg}
\graphicspath{ {./images/} }

\begin{document}

\pagestyle{headings}
\mainmatter

\title{Globally Continuous and Non-Markovian Crowd Activity Analysis from Videos}

\titlerunning{Globally Continuous and Non-Markovian Activity Analysis from Videos}

\authorrunning{He Wang and Carol O'Sullivan}

\author{He Wang\inst{1,2} and Carol O'Sullivan\inst{1,3}\thanks{Correspondence Author}}
\institute{Disney Research Los Angeles\thanks{This work is mostly done by the authors when they were with Disney Research Los Angeles.}, United States of America
\and
University of Leeds, United Kingdoms \\
\email{realcrane@gmail.com}
\and 
Trinity College Dublin, Ireland \\
\email{carol.osullivan@scss.tcd.ie}}
\maketitle

\begin{abstract}
Automatically recognizing activities in video is a classic problem in vision and helps to understand behaviors, describe scenes and detect anomalies. We propose an unsupervised method for such purposes. Given video data, we discover recurring activity patterns that appear, peak, wane and disappear over time. By using non-parametric Bayesian methods, we learn coupled spatial and temporal patterns with minimum prior knowledge. To model the temporal changes of patterns, previous works compute Markovian progressions or locally continuous motifs whereas we model time in a globally continuous and non-Markovian way. Visually, the patterns depict flows of major activities. Temporally, each pattern has its own unique appearance-disappearance cycles.  To compute compact pattern representations, we also propose a hybrid sampling method. By combining these patterns with detailed environment information, we interpret the semantics of activities and report anomalies. Also, our method fits data better and detects anomalies that were difficult to detect previously.
\end{abstract}

\section{Introduction}
Understanding crowd activities from videos has been a goal in many areas \cite{zhou_crowd_2010}. In computer vision, a number of subtopics have been studied extensively, including flow estimation \cite{ali_floor_2008}, behavior tracking \cite{antonini_behavioral_2006} and activity detection \cite{emonet_extracting_2011,zhou_learning_2014}. The main problem is essentially mining recurrent patterns over time from video data. In this work, we are particularly interested in mining recurrent spatio-temporal activity patterns, i.e., recurrent motions such as pedestrians walking or cars driving. Discovering these patterns can be useful for applications such as scene summarization, event counting or unusual activity detection. On a higher level, such patterns could be used to reduce the dimensionality of the scene description for other research questions.

Pattern finding has been previously addressed \cite{wang_trajectory_2008,wang_unsupervised_2009,emonet_extracting_2011}, but only either for the spatial case, a Markovian progression or local motifs. To consider temporal information in a global non-Markovian fashion, we propose a Spatio-temporal Hierarchical Dirichlet Process (STHDP) model. STHDP leverages the power of Hierarchical Dirichlet Process (HDP) models to cluster location-velocity pairs and time simultaneously by introducing two mutually-influential HDPs. The results are presented as activity patterns and their time-varying presence (e.g. appear, peak, wane and disappear).

Combined with environment information, our approach provides enriched information for activity analysis by automatically answering questions (such as what, where, when and how important/frequent) for each activity, which facilitates activity-level and higher-level analysis. The novelty and contributions of our work are as follows:
\begin{enumerate}
\item We present an unsupervised method for activity analysis that requires no prior knowledge about the crowd dynamics, user labeling or predefined pattern numbers.
\item Compared to static HDP variants, we explicitly model the time-varying presence of activity patterns. 
\item Complementary to other dynamic HDP variants, we model time in a globally continuous and non-Markovian way, which provides a new perspective for temporal analysis of activities. 
\item We also propose a non-trivial split-merge strategy combined with Gibbs sampling to make the patterns more compact.
\end{enumerate}

\subsection{Related Work}
Activities can be computed from different perspectives. On an individual level, tracking-based methods \cite{stauffer_learning_2000,oliver_bayesian_2000} and those with labeled motion features \cite{zelnik-manor_event-based_2001,zhong_detecting_2004} have been successful. On a larger scale, flow fields \cite{ali_floor_2008,lin_learning_2009} can be computed and segmented to extract meaningful crowd flows. However, these methods do not reveal the latent structures of the data at the flow level well where trajectory-based approaches prove to be very useful \cite{zhou_learning_2014,yi_understanding_2015,kitani_activity_2012,xie_inferring_2013}. Trajectories can be clustered based on dynamics \cite{zhou_learning_2014}, underlying decision-making processes \cite{kitani_activity_2012} or the environment \cite{xie_inferring_2013,yi_understanding_2015}. However, these works need assumptions or prior knowledge of the crowd dynamics or environment. Another category of trajectory-based approaches is unsupervised clustering to reveal latent structures \cite{wang_unsupervised_2009,emonet_extracting_2011,wang_trajectory_2011,varadarajan_sequential_2012}. This kind of approaches assumes minimal prior knowledge about the environment or cluster number. Our method falls into this category.

Non-parametric Bayesian models have been used for clustering trajectories. Compared to the methods mentioned above, non-parametric Bayesian models have been proven successful due to minimal requirements of prior knowledge such as cluster numbers and have thus been widely used for scene classifications \cite{fei-fei_bayesian_2005,sudderth_describing_2007}, object recognition \cite{sivic_discovering_2005}, human action detection \cite{niebles_unsupervised_2008} and video analysis \cite{kaufman_finding_2005,wang_unsupervised_2009}.  Initial efforts on using these kinds of models to cluster trajectories mainly focused on the spatial data \cite{wang_unsupervised_2009}. Later on, more dynamic models have been proposed \cite{wang_trajectory_2011,emonet_extracting_2011,varadarajan_sequential_2012}. Wang et al. \cite{wang_trajectory_2011} propose a dynamic Dual-HDP model by assuming a Markovian progression of the activities and manually sliced the data into equal-length intervals. Emonet et al. \cite{emonet_extracting_2011,emonet_temporal_2014} and Varadarajan et al.\cite{varadarajan_sequential_2012} model time as part of local spatio-temporal patterns, but no pattern progression is modeled.  The former requires manual segmentation of the data and assumes the Markovian property, which does not always apply and could adversely affect detecting temporal anomalies. The latter focuses on local continuity in time and cannot learn time activities well when chunks of data are missing.

Inspired by many works in Natural Language Processing and Machine Learning \cite{teh_hierarchical_2006,dubey_non-parametric_2012,wang_split-merge_2012,lin_construction_2010,blei_distance_2011,wang_topics_2006}, we propose a method that is complementary to the methods above in that we model time in a globally continuous and non-Markovian way. We thus avoid manual segmentation and expose the time-varying presence of each activity. We show how our method fits data better and in general more aligned with human judgments. In addition, our method is good at detecting temporal anomalies that could be missed by previous methods.
\section{Methodology}
\subsection{Spatio-temporal Hierarchical Dirichlet Processes}
Given a video, raw trajectories can be automatically estimated by a standard tracker and clustered to show activities, with each activity represented by a trajectory cluster.  One has the option of grouping trajectories in an unsupervised fashion where a distance metric needs to be defined, which is difficult due to the ambiguity of the associations between trajectories and activities across different scenarios. Another possibility is to cluster the individual observations of trajectories, such as locations, in every frame.  Since observations of the same activity are more likely to co-occur, clustering co-occurring individual observations will eventually cluster their trajectories. This problem is usually converted into a data association problem, where each individual observation is associated with an activity. However, it is hard to know the number of activities in advance, so Dirichlet Processes (DPs) are used to model potentially infinite number of activities. In this way, each observation is associated with an activity and trajectories can be clustered based on a \textit{softmax} scheme (a trajectory is assigned to the activity that gives the best likelihood on its individual observations). During the data association, DPs also automatically compute the ideal number of activities so that the co-occurrences of observations in the whole dataset can be best explained by an finite number of activities. To further capture the commonalities among the activities across different data segments, Hierarchical DPs (HDPs) are used, where one DP captures the activities in one data segment and another DP on a higher level captures all possible activities.

\begin{figure}
\centering
  \begin{minipage}[b]{0.4\textwidth}
    \includegraphics[width=\textwidth]{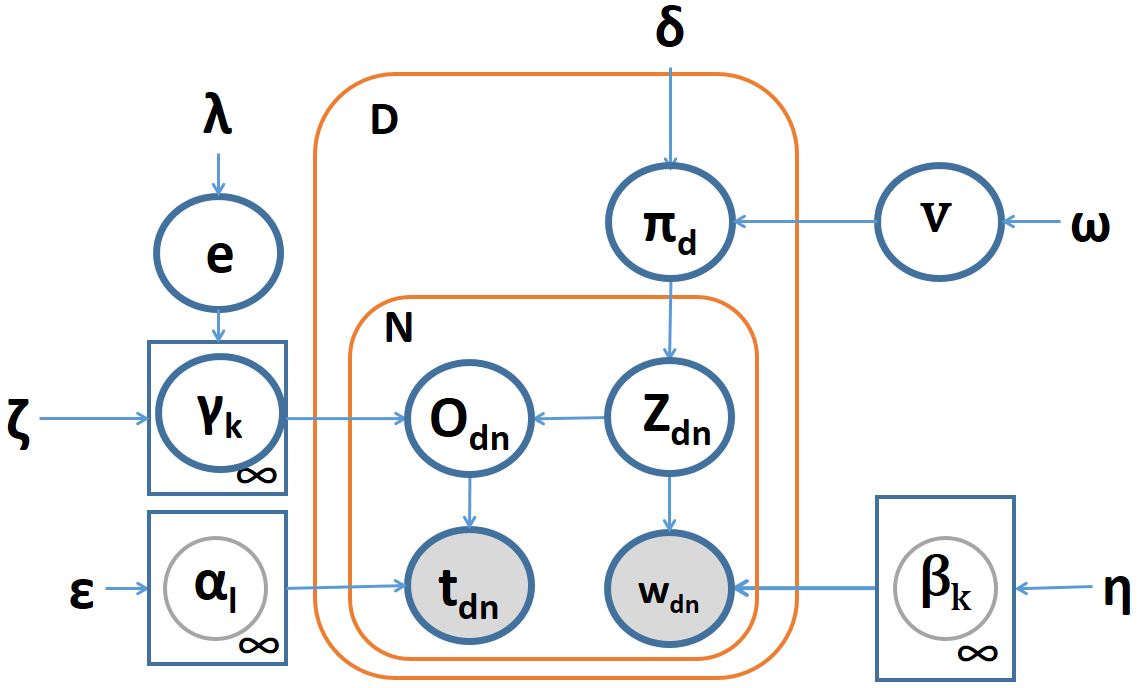}
    \caption{STHDP model}
    \label{fig:STHDP}
  \end{minipage}
  \begin{minipage}[b]{0.5\textwidth}
    \includegraphics[width=\textwidth]{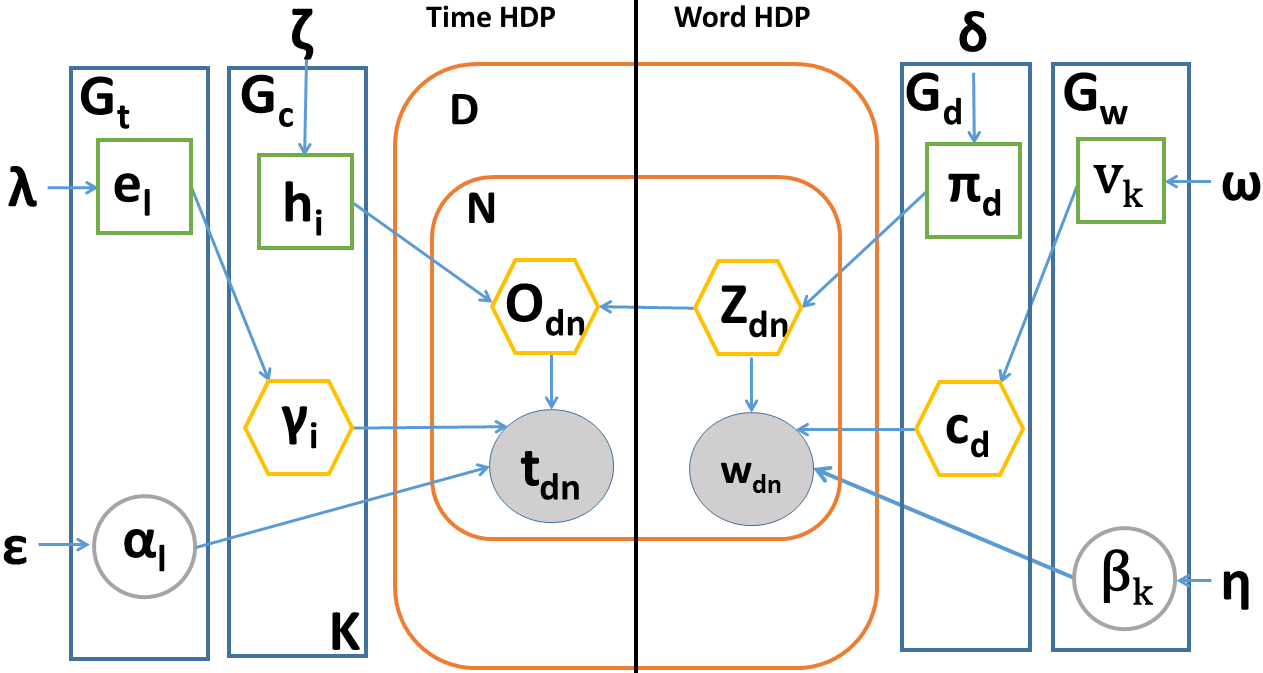}
    \caption{Model used for sampling.}
    \label{fig:STHDPSampling}
  \end{minipage}

\end{figure}

To cluster video data in the scheme explained above, we discretize the camera image into grids, that discretizing a trajectory into locations. We also discretize the velocity into several subdomains based on the orientation so that each location also comes with a velocity component. Finally, we can model activities as Multinomial distributions of time-stamped location-velocity pairs $\{w, t\}$, $w$ = ($p_x$, $p_y$, $p'_x$, $p'_y$) where ($p_x$, $p_y$) is the position, ($p'_x$, $p'_y$) is the velocity and each $\{w, t\}$ is an \textit{observation}.  Given multiple data segments consisting of location-velocity pairs, we can use the HDP scheme explained above to cluster trajectories. In addition, our STHDP also has a temporal part. Consider that a time data segment is formed by all the time stamps of the observations associated with one activity, then the distribution of these time stamps reflect the temporal changes of the activity. Since these time stamps might come from different periods (e.g. an activity appears/disappears multiple times), we need a multi-modal model to capture it. Again, since we do not know how many periods there are, we can use a DP to model this unknown too, which can be captured by an infinite mixture of Gaussians over the time stamps. Finally, to compute the time activities across different time data segments, we also use a HDP to model time. The whole scheme is explained by a Bayesian model shown in \figref{STHDP}.

To mathematically explain our model, we first introduce some background and terminologies. In a \textit{stick-breaking} representation \cite{sethuraman_constructive_1994} of a DP: $G = \sum_{k=1}^{\infty} \sigma_k(v) \beta_k$, where $\sigma_k(v)$ is \textit{iteratively} generated from $\sigma_k(v) = v_k\prod^{k-1}_{j=1}(1-v_j)$, $\sum^{\infty}_{k=1} \sigma_k(v) = 1$, $v \sim Beta(1, \omega)$ and $\beta_k \sim H(\eta)$. $\beta_k$ are DP $\textit{atoms}$ drawn from some base distribution H and $\sigma_k(v)$ are \textit{stick proportions}. We refer to the iterative generation of sticks $\sigma_k(v)$ from $v$ as $\sigma \sim GEM(v)$, as in \cite{teh_hierarchical_2006}. Following the convention of topics models, we refer to a location-velocity pair as a \textbf{\textit{word}}, its time stamp as a \textbf{\textit{time\ word}}, activity patterns as \textbf{\textit{word\ topics}} and time activities as \textbf{\textit{time topics}}. A data segment is called a \textbf{\textit{document}} and a time stamp data segment is called a \textbf{\textit{time\ document}}. The whole dataset is called a  \textbf{\textit{corpus}}. The overall activities and time activities we are aiming for are the corpus-level word topics and time topics. 

\figref{STHDP} depicts two HDPs: a word HDP and a time HDP, respectively modeling the spatial and temporal data as described above. The word HDP starts with a DP over corpus-level word topics $v \sim GEM(\omega)$.  In each document, there exists a DP $\pi_d \sim DP(v, \sigma)$ governing the document-level topics. For each word, a topic indicator is sampled by $Z_{d_n} \sim \pi_d$ and the word is generated from $w_{dn} | \beta_{Z_{d_n}} \sim Mult(\beta_{Z_{d_n}})$. The time HDP models how word topics evolve. Unlike previous models, it captures two aspects of time: continuity and multi-modality. Continuity is straightforward. Multi-modality means a word topic can appear/disappear several times. Imagine all the time words associated with the words under one word topic. The word topic could peak multiple times which means its time words are mainly aggregated within a number of time intervals. Meanwhile, there can be infinitely many word topics and some of their time words share some time intervals. Finally, there can be infinitely many such shared time intervals or time topics, which are modeled by an infinite mixture of Gaussians, of which each component is a time topic. A global DP $e \sim GEM(\lambda)$ governs all possible time topics. Then, for each corpus-level word topic $k$, a time DP $\gamma_k \sim DP(\zeta, e)$ is drawn. Finally, when a specific time word is needed, its $Z_{d_n}$ indicates its word topic based on which we draw a time word indicator $O_{d_n} \sim \gamma_{Z_{d_n}}$ and a time word is generated from $t_{dn} | \alpha_{O_{d_n}} \sim Normal(\alpha_{O_{d_n}})$. In this way, each word topic corresponds to a subset of time topics with different weights. Thus a Gaussian Mixture Model (GMM) is naturally used for every word topic. Due to the space limit, the generative process of \figref{STHDP} is explained in the supplementary material.


\subsection{Posterior by Sampling}
To compute the word and time topics we need to compute the posterior of STHDP. Both sampling and variational inference have been used for computing the posterior of hierarchical models \cite{teh_hierarchical_2006,hoffman_stochastic_2013}. After our first attempt at variational inference, we found that it suffers from the sub-optimal local minima because of the word-level coupling between HDPs. Naturally, we resort to sampling. Many sampling algorithms have been proposed for such purposes \cite{dubey_non-parametric_2012,chang_parallel_2014,hughes_effective_2012}. However, due to the structure of STHDP, we found that it is difficult to derive a parallel sampling scheme such as the one in \cite{chang_parallel_2014}. Finally, we employ a hybrid approach that combines Gibbs and Metropolis-Hasting (MH) sampling based on the stick-breaking model shown in \figref{STHDPSampling}, the latter being the split-merge (SM) operation. For Gibbs sampling, we use both Chinese Restaurant Franchise (CRF) \cite{teh_hierarchical_2006} and modified Chinese Restaurant Franchise (mCRF) \cite{dubey_non-parametric_2012}. As there are two HDPs in STHDP, we fix the word HDP when sampling the time HDP which is a standard two-level HDP, so we run CRF sampling \cite{teh_hierarchical_2006} on it. For the word HDP, we run mCRF. Please refer to the supplementary material for details.

HDPs suffer from difficulties when two topics are similar, as the sampling needs to go through a low probability area to merge them \cite{jain_split-merge_2000}. This is particular problematic in our case because each observation is pulled by two HDPs. Split-merge (SM) methods have been proposed \cite{jain_split-merge_2000,dahl_sequentially-allocated_2005} for Dirichlet Processes Mixture Models, but they do not handle HDPs. Wang et al. \cite{wang_split-merge_2012} proposes an SM method for HDP, but only for one HDP, whereas STHDP has two entwined HDPs.  We propose a Metropolis-Hasting (MH) sampling scheme to perform SM operations. In our version of the CRF metaphor, word topics and time topics are called \textbf{\textit{word\ dishes}} and \textbf{\textit{time\ dishes}}. Word documents are called \textbf{\textit{restaurants}} and time documents are called \textbf{\textit{time\ restaurants}}. Some variables are given in \tabref{crfVarShort}. Similar to \cite{wang_split-merge_2012}, we also only do split-merge on the word dish level. We start with the SM operations for the word HDP. In each operation, we randomly choose two word tables, indexed by i and j. If they serve the same dish, we try to split this dish into two, and otherwise merge these two dishes. Since the merge is just the opposite operation of split, we only explain the split strategy here.

\begin{table}
\centering
\caption{Variables in CRF}
\label{tab:crfVarShort}
\begin{tabular}{|p{1cm}|p{9cm}|}
\hline
$v_w$ & a word in the vocabulary \\
\hline
$V_w$ & the size of the vocabulary \\
\hline
$n_{jik}$ & the number of words in restaurant j at table i serving dish k\\
\hline
$z_{ji}$ & the table indicator of the $i$th word in restaurant j\\
\hline
$m_{jk}$ & the number of word tables in restaurant j serving dish k\\
\hline
$m_{j\cdot}$ & the number of word tables in restaurant j\\
\hline
$K$ & the number of word dishes\\
\hline
\end{tabular}
\end{table}

Following \cite{jain_split-merge_2000}, the MH sampling acceptance ratio is computed by:
\begin{equation}
\label{eq:MH}
a(c^*, c) = min \{1, \frac{q(c|c^*)}{q(c^*|c)}\frac{P(c^*)}{P(c)}\frac{L(c^*|y)}{L(c|y)}\}
\end{equation}
where $c^*$ and $c$ are states (table and dish indicators) after and before split and $q(c^*|c)$ is the split transition probability. The merge transition probability $q(c|c^*)$ = 1 because there is only one way to merge. $P$ is the prior probability, $y$ are the observations, so $L(c^*|y)$ and $L(c|y)$ are the likelihoods of the two states.  The split process of MH is: sample a dish, split it into two according to some process, and compute the acceptance probability $a(c^*, c)$. Finally, sample a probability $\phi \sim Uniform(0, 1)$. If $\phi > a(c^*, c)$, it is accepted, and rejected otherwise. The whole process is done only within the sampled dish and two new dishes. All the remaining variables are fixed.

Now we derive every term in \eqref{MH}. The state $c$ consists of the table and dish indicators. Because the time HDP needs to be considered when sampling the word HDP, the prior of table indicators is:

\begin{equation}
\label{eq:tablePrior}
p(\textbf{z}_j) = \frac{\delta^{m_{j\cdot}} \prod^{m_{j\cdot}}_{t=1}(n_{jt} p(t|\bullet) - 1)!}{\prod_{i=1}^{n_{j\cdot}}(i+\delta-1)}
\end{equation}
where $p(t|\bullet)$ represents the marginal likelihood of all time words involved. Similarly, for word dish indicators:
\begin{equation}
\label{eq:dishPrior}
p(\textbf{k}) = \frac{\omega^{K} \prod^{K}_{k=1}(m_{\cdot k} p(t|\bullet) - 1)!}{\prod_{i=1}^{m_{\cdot\cdot}}(i+\omega-1)}
\end{equation}
Now we have the prior for p(c):
\begin{equation}
\label{eq:statePrior}
p(c) = p(\textbf{k})\prod^D_{j=1}p(\textbf{z}_j)
\end{equation}
where $D$ is the number of restaurants; $p(c^*)$ can be similarly computed.

Now we derive $q(c^*|c)$. Assume that tables i and j both serve dish k. We denote $S$ as the set of indices of all tables also serving dish k excluding i and j. In the split state, k is split into $k_1$ and $k_2$. We denote $S_1$ and $S_2$ as the sets of indices of tables serving dishes $k_1$ and $k_2$. We first assign table i to $k_1$ and j to $k_2$, then allocate all tables indexed by $S$ into either $k_1$ or $k_2$ by \textit{sequential\ allocation\ restricted\ Gibbs\ sampling} \cite{dahl_sequentially-allocated_2005}:

\begin{equation}
\label{eq:sequentialAlloc}
p(SK = k_j | S_1, S_2) \propto m_{\cdot k_j} f_{k_j}(\textbf{w}_{SK})p(\textbf{t}_{SK} | \bullet)
\end{equation}
where j = 1 or 2, $SK \in S$, $\textbf{w}_{SK}$ is all the words at table $SK$ and $m_{\cdot k_j}$ is the total number of tables assigned to $k_j$. All the tables in $S$ are assigned to either $k_1$ or $k_2$. We still approximate $p(\textbf{t}_{SK} | \bullet)$ by $\hat{p}(\textbf{t}_{SK}|\bullet)$ as we do for Gibbs sampling (cf. supplementary material). Note that during the process, the sizes of S1 and S2 constantly change. Finally, we compute $q(c^*|c)$ by \eqref{qsplit}:

\begin{equation}
\label{eq:qsplit}
q(c^*|c) = \prod_{i \in S}p(k^i = k | S1, S2)
\end{equation}
Finally, the likelihoods are:

\begin{equation}
\label{eq:liks}
\frac{L(c^*|y)}{L(c|y)} = \frac{f^{lik}_{k_1}(\textbf{w}_{k_1}, \textbf{t}_{k_1} | c^*) f^{lik}_{k_2}(\textbf{w}_{k_2}, \textbf{t}_{k_1}| c^*)}{f^{lik}_k(\textbf{w}_{k}, \textbf{t}_{k} | c)}
\end{equation}
where 
\begin{equation}
\label{eq:condLik}
f^{lik}(w, t | c) = \frac{\Gamma(V_w\eta)}{\Gamma(n_{\cdot\cdot k} + V_w\eta)}\frac{\prod_{v_w}\Gamma(n_{\cdot\cdot k}^{v_w} + \eta)}{\Gamma^{V_w}(\eta)}p(t | \bullet)
\end{equation}
$\Gamma$ is the gamma function, $n_{\cdot\cdot k}$ is the number of words in topic k, $n_{\cdot\cdot k}^{v_w}$ is the number of words $v_w$ assigned to topic k, and $p(t | \bullet)$ is the likelihood of the time words involved.

Now we have fully defined our split-merge operations. Whenever an SM operation is executed, the time HDP needs to be updated. During the experiments, we do one iteration of SM after a certain number of Gibbs sampling iterations.

We found it is unnecessary to do SM on the time HDP for two reasons. First, we already implicitly consider the time HDP here through Equations \ref{eq:tablePrior} - \ref{eq:condLik} and an SM operation on the word HDP will affects the time HDP. Second, we want the word HDP to be the dominating force over the time HDP in SM.  A merge operation on the word topics will cause a merge operation on the time HDP, which makes the patterns more compact. The reverse is not ideal because it can merge different word topics. However, this does not mean that the time HDP is always dominated. Its impact in the Gibbs sampling plays a strong role in clustering samples that are temporally close together while separating samples that are not.

\section{Experiments}
Empirically, we use a standard set of parameters for all our experiments. The prior $Dirichlet$($\eta$) is a symmetric Dirichlet where $\eta$ is initialized to 0.5. For all $GEM$ weights, we put a vague $Gamma$ prior, $Gamma(0.1, 0.1)$, on their $Beta$ distribution parameters, which are updated in the same way as \cite{teh_hierarchical_2006}. The last is the Normal-Inverse-Gamma prior, $NIG(\mu, \lambda, \sigma_1, \sigma_2)$, where $\mu$ is the mean, $\lambda$ is the variance scalar, and $\sigma_1$ and $\sigma_2$ are the shape and scale. For our datasets, we set $\mu$ to the sample mean, $\lambda$ = 0.01, $\sigma_1$ = 0.3 and $\sigma_2 = 1$. Because both the Gamma and NIG priors are very vague here, we find that the performance is not much affected by different values, so we fix the NIG parameters. For simplicity, we henceforth refer to all activity Patterns with the letter P.

\subsection{Synthetic Data}

\begin{figure}
\centering
\includegraphics[scale=0.27]{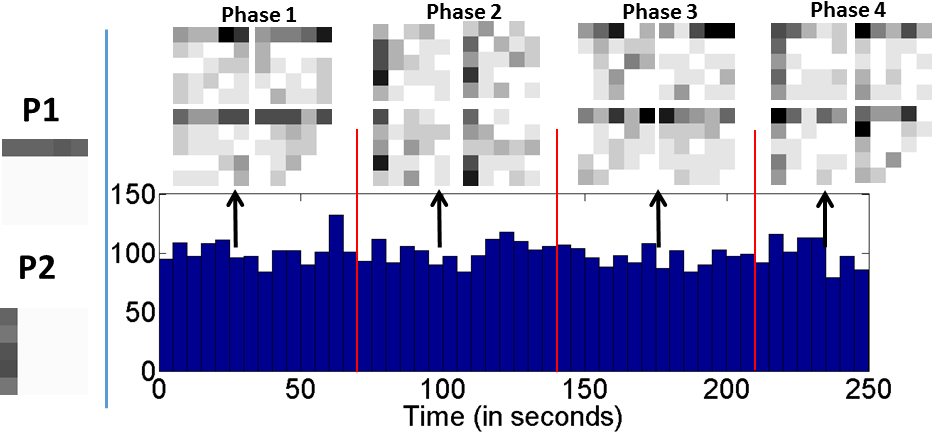}
\includegraphics[scale=0.27]{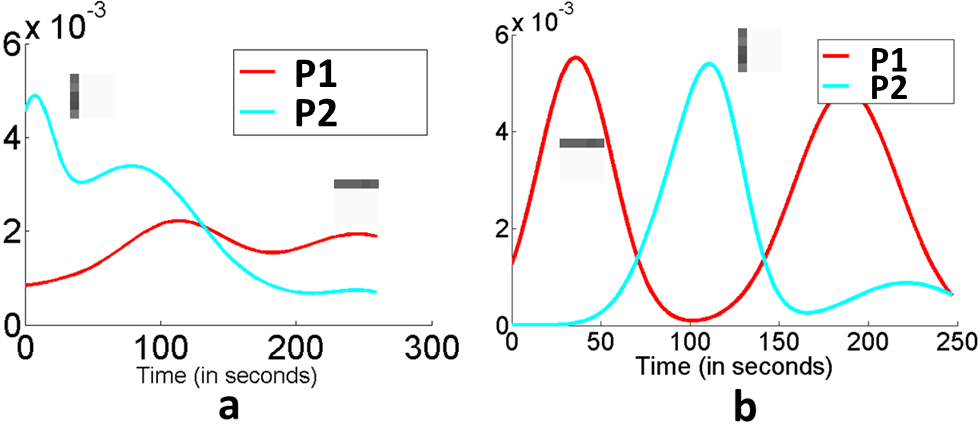}
\caption{Left: Two ground truth patterns on a 5 $\times$ 5 grid (one distributed over a horizontal bar, the other over a vertical bar) and generated data over four time periods (P1 is used for generating data in phase 1 and phase 3, P2 is used for data in phase 2 and both are used for phase 4. Some document examples and the histogram of observation numbers are shown for each phase). Right: Learned patterns and their time activities by (a) HDP \cite{teh_hierarchical_2006} and (b) STHDP.}
\label{fig:grid}
\end{figure}

A straightforward way to show the effectiveness of our method is to use synthetic data where we know the ground truth so that we can compare learned results with the ground truth. Similar to \cite{wang_unsupervised_2009}, we use grid image data where the ground truth patterns  (\figref{grid} P1 and P2) are used to generate synthetic document data. Different from \cite{wang_unsupervised_2009}, to show the ability of our model in capturing temporal information, we generate data for 4 periods where the two ground truth patterns are combined in different ways for every period. Documents are randomly sampled from the combined pattern for each period. \figref{grid} Right shows the learned patterns and their time activities from HDP \cite{teh_hierarchical_2006} and STHDP. The time activities for HDP are represented by a GMM over the time words associated with the activity pattern. For STHDP, a GMM naturally arises for each pattern by combining the time patterns and their respective weights.

Both HDP and STHDP learn the correct spatial patterns. However, HDP assumes exchangeability of all data points thus its time information is not meaningful. In contrast, STHDP not only learns the correct patterns, but also learns a multi-modal representation of its temporal information which reveals three types of information. First, all GMMs are scaled proportionally to the number of associated data samples, so their integrals indicate their relative importance. In \figref{grid} Left, the number of data samples generated from P1 is roughly twice as big as that from P2. This is reflected in \figref{grid} Right (b) (The area under the red curve is roughly twice as big as that under the blue curve). Second, each activity pattern has its own GMM to show its presence over time. The small bump of the blue curve in (b) shows that there is a relatively small number of data samples from P2 beyond the $210^{th}$ second. It is how we generated data for phase 4. Finally, different activity patterns have different weights across all the time topics. Conversely, at any time, the data can be explained by a weighted combination of all activity patterns. Our method provides an enriched temporal model that can be used for analysis in many ways. 

\subsection{Real Data}
In this section, we test our model on the Edinburgh dataset \cite{majecka_statistical_2009}, the MIT Carpark database \cite{wang_trajectory_2011} and New York Central Terminal \cite{yi_understanding_2015}, referred to as \textbf{Forum}, \textbf{Carpark} and \textbf{TrainStation} respectively. They are widely used to test activity analysis methods \cite{wang_trajectory_2011,yi_understanding_2015,wang_unsupervised_2009,luber_socially-aware_2012,almingol_learning_2013}. Each dataset demonstrates different strengths of our method. Forum consists of indoor video data with the environment information available for semantic interpretation of the activities.  Carpark is an outdoor scene consisting of periodic video data that serves as a good example to show the multi-modality of our time modeling. TrainStation is a good example of large scenes with complex traffic. All patterns are shown by representative (high probability) trajectories.

\subsubsection{Forum Dataset}
\begin{figure}
\centering
\includegraphics[scale=0.4]{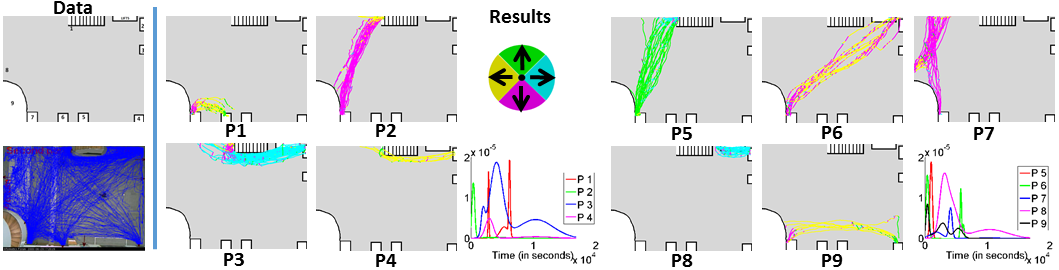}
\caption{Top Left: Environment of Edinburgh dataset. Bottom Left: Trajectories overlaid on the environment. Right: Some activities shown by representative trajectories and their respective time activities. Colors indicate orientations described by the legend in the middle.}
\label{fig:forumPatterns}
\end{figure}

The forum dataset is recorded by a bird's eye static camera installed on the ceiling above an open area in a school building. 664 trajectories have been extracted as described in \cite{majecka_statistical_2009}, starting from 14:19:28 GMT, 24 August 2009 and lasting for 4.68 hours. The detailed environment is shown in \figref{forumPatterns} (left). We discretize the 640 * 480 camera image into 50$\times$50 pixel grids and the velocity direction into 4 cardinal subdomains and then run a burn-in 50 iterations of Gibbs sampling. For the first 500 iterations, MH sampling is done after every 10 Gibbs sampling iterations. Then we continue to run it for another 1500 iterations. 


Nine patterns are shown in \figref{forumPatterns}, where the semantics can be derived by also considering the environment information in the form of Zones Z1: Stairs, Z2-Z7: Doors, Z8: Conference room, Z9: a floating staircase that blocks the camera view but has no semantic effect here. P3 and P4 are two groups of opposite trajectories connecting Z1 and Z2. We observe many more trajectories in P3 than P4. From the detailed environment information, we know that the side door outside of Z2 is a security door. This door can be opened from the inside, but people need to swipe their cards to open it from outside, which could explain why there are more exiting than entering activities through Z2. P2 is the major flow when people come down the stairs and go to the front entrance. P1 has a relatively small number of trajectories from Z6 to Z7, i.e., leaving through the front entrance. From the temporal point of view, the two major incoming flows can be seen in \figref{forumPatterns} P4 and \figref{forumPatterns} P5, spanning the first half of the data. We also spot a pattern with a high peak at the beginning (around 2:34pm), shown by \figref{forumPatterns} P7, which connects the second part of the area and the conference room. We therefore speculate that there may have been a big meeting around that time.

\subsubsection{Carpark Dataset}
The Carpark dataset was recorded by a far-distance static camera over a week and 1000 trajectories were randomly sampled as shown in \figref{carparkPatterns} Left. Since this dataset is periodic, it demonstrates the multi-modality of our time modeling. We run the sampling in the same way as in the Forum experiment.

\InsertFig{\includegraphics[scale=0.4]{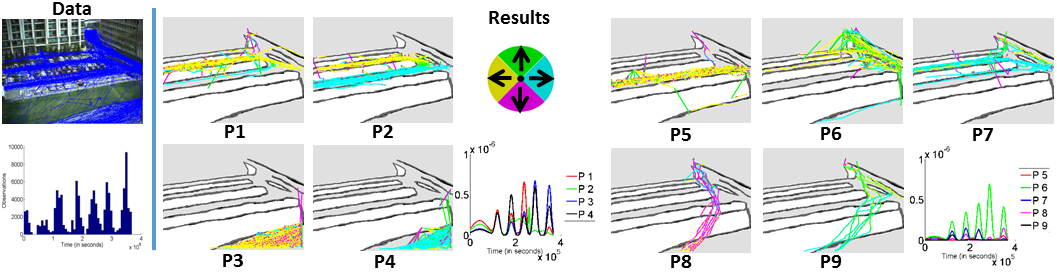}}{Top Left: Environment of the car park. Bottom Left: Observation numbers over time. Right: Some activities shown by representative trajectories and their respective time activities. Colors indicate orientations described by the legend in the middle.}{carparkPatterns}{th}

Four top activity patterns and their respective time presence are shown in \figref{carparkPatterns}. P1 is the major flow of in-coming cars, P2 is an out-going flow, and P3 and P4 are two opposite flows. Unfortunately, we do not have detailed environment information as we do from Forum for further semantic interpretations. The temporal information shows how all peaks are captured by our method, but different patterns have different weights in different periods. 



\subsubsection{TrainStation Dataset}
The TrainStation dataset was recorded from a large environment and 1000 trajectories were randomly selected for the experiment. The data and activities are shown in \figref{trainStationPatterns}.

\InsertFig{\includegraphics[scale=0.4]{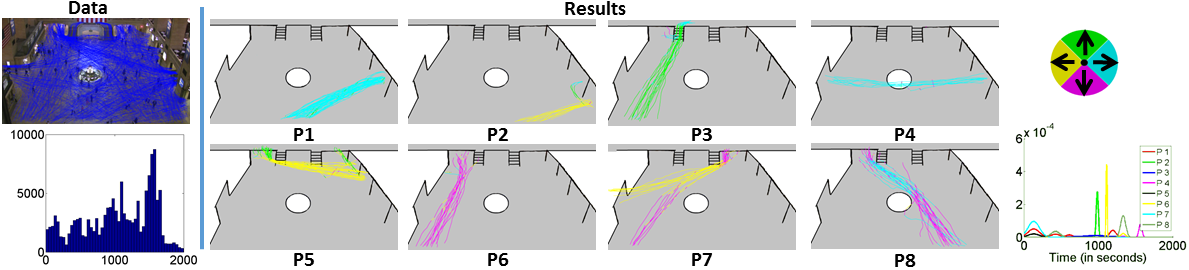}}{Top Left: Environment of the New York Central Terminal. Bottom Left: Observation numbers over time. Right: Some activities shown by representative trajectories and their respective time activities. Colors indicate orientations described by the legend on the right.}{trainStationPatterns}{th}

\subsection{Split and Merge}


We test the effectiveness of split-merge (SM) by the per-word log likelihood by:
\begin{eqnarray}
P_{per-word} = \frac{\sum_{n = 1}^NP(w_n, t_n | \beta, v, \alpha, e)}{N} = \nonumber \\
\frac{\sum_{n = 1}^N(\sum_{k=1}^KP(w_n | \beta_k, v_k)\sum_{l=1}^LP(t_n | \alpha_l, e_l, \bullet))}{N}
\label{eq:lik}
\end{eqnarray}
where $N$, $K$ and $L$ are the number of observations, learned spatial activities and time activities respectively. $\beta$ and $v$ are the spatial activities and their weights, $\alpha$ and $e$ are the time activities and their weights, $\bullet$ represents all the other factors. In general, we found that SM increases the likelihood thus improves the model fitness on the data. Also, we found that MH sampling is more likely to pick a merge operation than a split. One reason is the time HDP tends to separate data samples that are temporally far away from each other, thus causing similar patterns to appear at different times. A merge on those patterns has higher probability, thus is more likely to be chosen. Merging such spatially similar patterns makes each final activity unique. It is very important because not only does it make the activities more compact, it also makes sure that all the time activities for a particular spatial activity can be summarized under one pattern.


%
%
%
\subsection{Anomaly Detection}
\label{sec:anomaly}
For anomaly detection, \figref{outliers} shows the top outliers (i.e., unusual activities) in three datasets: (g) and (l) show trajectories crossing the lawn, which is rare in our sampled data; (i) shows a trajectory of leaving the lot then returning. In the latter case, the trajectory exited in the upper lane, whereas most activities involve entering in the upper lane and exiting in the bottom lane. The outliers in the Forum are also interesting. \figref{outliers} (a) shows a person entering through the front entrance, checking with the reception then going to the conference room; (b) shows a person entering through Z2 then leaving again;(d) is unexpected because visually it should be in \figref{forumPatterns} (P7), but we found that the pattern peaks around 2:34pm and falls off quickly to a low probability area before 2:27:30pm whereas \figref{outliers} (c) occurs between 2:26:48pm-2:26:53pm. This example also demonstrates that our model identifies outliers not only on the spatial domain but also on the time domain. We also found similar cases in \figref{outliers} (k), (l) and (o) that are normal when only looking at the spatial activities but become anomalies when the timing is also considered.

\InsertFig{\includegraphics[scale=0.5]{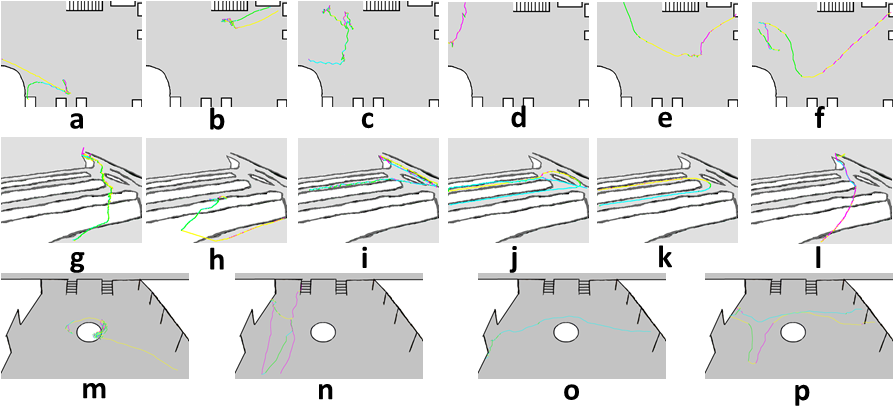}}{Top: Outliers in Forum. Middle: Outliers in Carpark. Bottom: Outliers in TrainStation.}{outliers}{th}

\subsection{Comparison}

\subsubsection{Qualitative Comparison}
Our model is complementary to dynamic non-parametric models such as DHDP \cite{wang_trajectory_2011} and MOTIF \cite{emonet_extracting_2011}. Theoretically, our time modeling differs in two aspects: continuity and multi-modality. Wang et.al \cite{wang_trajectory_2011} manually segment data into equal-length episodes and Emonet et.al \cite{emonet_extracting_2011} model local spatio-temporal patterns. Our method treats time as a globally continuous variable. Different time modeling affects the clustering differently. All three models are variants of HDP which assumes the \textit{exchangeability} of data samples. This assumption is overly strict when time is involved because it requires any two samples from different time to be exchangeable. The manual segmentation \cite{wang_trajectory_2011} restricts the exchangeability within segment. Enforcing a peak-and-fall-off time prior \cite{emonet_extracting_2011} has similar effects. 


For DHDP, we segment both datasets equally into 4 episodes. We run it for 5000 iterations on each episode. For MOTIF, we used the author's implementation \cite{emonet_extracting_2011}. For parameter settings, we use our best effort at picking the same parameters for three models, e.g. the Dirichlet, DP, Gamma and Beta distribution parameters on each level. Other model-specific parameters are empirically set to achieve the best performance. Also, since there is no golden rule regarding when to stop the sampling, we use the time DHDP models takes and run the other two for roughly the same period of time.

\begin{figure}
\centering
\includegraphics[scale=0.2]{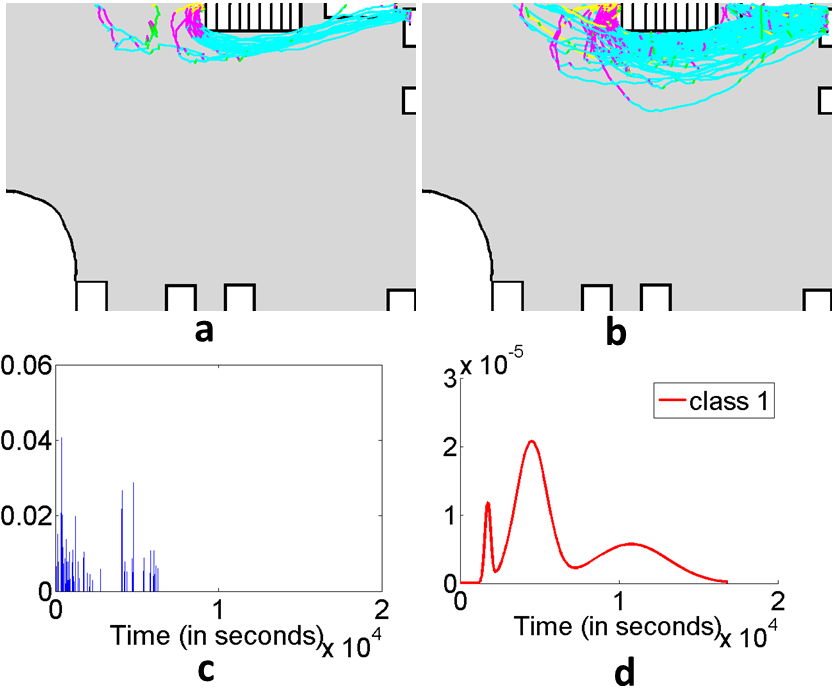}
\includegraphics[scale=0.2]{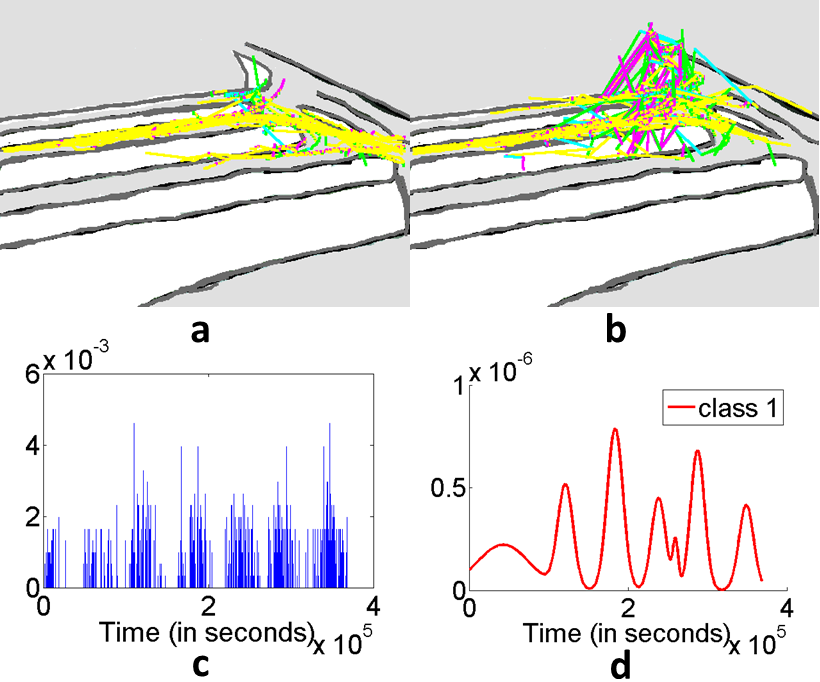}
\caption{Left: Forum dataset. (a) A pattern learned by DHDP. (b) A pattern learned by MOTIF. (c) The topic starting time distribution over time from MOTIF. (d) The time activities of \figref{forumPatterns} (P3) from STHDP. Right: Carpark dataset. (a) A pattern learned by DHDP. (b) A pattern learned by MOTIF. (c) The topic starting time distribution over time from MOTIF. (d) The time activities of \figref{carparkPatterns} (1) from STHDP.}
\label{fig:comparison}
\end{figure}

Since all three methods learn similar spatial activities, we mainly compare the temporal information. \figref{comparison} shows one common pattern found by all three methods. The temporal information of DHDP is simply the weight of this activity across different episodes. To get a dense distribution, smaller episodes are needed, but the ideal size is not clear. Therefore, we only plot the temporal information for MOTIF and STHDP.  In \figref{comparison} Left, (c) is the starting time probability distribution of \figref{comparison} Left (b). The distribution is discrete and shows how likely it is that this pattern could start at a certain time instance, which reports quite different information from our pattern. \figref{comparison} Left (d) shows the time activities of \figref{forumPatterns} (P3), which is continuous and shows its appearance, crescendo, multiple peaks, wane and disappearance. An interesting fact is that both methods capture this pattern within the first 8000 seconds while our model also captures a small bump beyond the first 8000 seconds. By looking at the data, we find that there are indeed a few trajectories belonging to this pattern beyond the first 8000 seconds.  \figref{comparison} Right shows a common pattern in the Carpark dataset. Both MOTIF and STHDP capture the periodicity as seen in P3 and P4. They mainly differ at the start in that P3 captures two peaks whereas P4 captures one. Note that the two peaks that P3 captures depict how likely it is that the activity starts at those time instances, while STHDP captures the time span of that activity, which is essentially the same.

\subsubsection{Quantitative Comparison}
Because all three methods model time differently, it is hard to do a fair quantitative comparison. As a general metric to evaluate a model's ability to predict, we use the per-word log likelihood (\eqref{lik}). We hold out 10\% of the data as testing data and evaluate the performance of the three models with respect to how well they can predict the testing data. We show the best per-word log likelihood of the methods after the burn-in period in \tabref{pwll}. 

\begin{table}[th]
\centering
\begin{tabular}{|c|c|c|c|}
\hline
  & STHDP & DHDP & MOTIF\\
 \hline
 Forum & -3.84 & -9.8 & -54.38\\
 \hline
 Carpark & -2.8 & -7.75 & -62.13\\
 \hline
 TrainStation & -3.5 & -4.9 & -62.2\\
 \hline
\end{tabular}
\begin{tabular}{|c|c|c|c|}
\hline
  $r_{correct}$/$r_{complete}$& STHDP & DHDP & MOTIF\\
 \hline
 Forum & 0.92/0.88 & 0.95/0.63 & 0.87/0.78\\
 \hline
 Carpark & 0.83/0.9 & 0.89/0.31 & 0.85/0.42\\
 \hline
 TrainStation & 0.84/0.75 & 0.72/0.55 & 0.69/0.58\\
 \hline
\end{tabular}
\caption{Left: Best per-word log likelihoods. Right: $r_{correct}$ and $r_{complete}$ accuracies from 0-1, 1 is the best.}
\label{tab:pwll}
\end{table}

This experiment favors DHDP and MOTIF. Because DHDP learns topics on different episodes, when computing the likelihood of a testing sample $\{w, t\}$, we only  weighted-sum its likelihoods across the topics learned within the corresponding episode. So the likelihood is $p(w|t, \bullet)$ instead of $p(w, t|\bullet)$ where $
\bullet$ represents all model parameters and the training data. For MOTIF, the learned results are topics as well as their durations in time. We compute the likelihood of a testing sample by averaging the likelihoods across all topics whose durations contains $t$, i.e., $p(w, t|\beta_k, t \in rt_k, \bullet)$ where $\beta_k$ is the topic with duration $rt_k$. For STHDP, the likelihood is computed across all word topics and all time topics, $p(w, t|\bullet)$, which is much more strict.  We found that STHDP outperforms both DHDP and MOTIF, with MOTIF performing more poorly than the other two. We initially found the results surprising given the fact that MOTIF learns similar spatial activities to the other two. Further investigations showed that, since the testing data is randomly selected, this causes gaps in the training data in time. As a consequence of the discrete nature of the time representation in MOTIF, all MOTIF topics have low probabilities in those gaps, thus causing the low likelihoods. Removing the time and only considering spatial activities in this case may help but would not be fair to the other two methods.

Next, we compute the correctness/completeness of the three methods as in \cite{wang_trajectory_2011}. Correctness is the accuracy of trajectories of different activities not clustered together while completeness is the accuracy of trajectories of the same activity clustered together. To get the ground truth data for each dataset, the trajectories were first roughly clustered into activities. Then 2000 pairs of trajectories were randomly selected where each pair comes from the same activity and another 2000 pairs were randomly selected where each pair comes from two different activities. Finally these 4000 pairs of trajectories for each dataset were labeled and compared with the results of our method. We denote the correctness as $r_{correct}$ and the completeness as $r_{complete}$. Because estimating the number of clusters is hard, it was only needed to judge whether a pair of trajectories was from the same activity or not. The correctness/completeness metric indicates that grouping all the trajectories in the same cluster results in 100\% completeness and 0\% correctness while putting every trajectory into a singleton cluster results in 100\% correctness and 0\% completeness. So only an accurate clustering can give good overall accuracies. \tabref{pwll} Right shows the accuracies. STHDP outperforms the other two on $r_{complete}$ across all datasets. Its $r_{correct}$ is higher in TrainStation and slightly worse in Forum and Carpark but the difference is small (within 6\%). 

Finally, we discuss how different temporal modeling could lead to different temporal anomaly detection. Most of the outliers in \figref{outliers} are \textit{spatial} outliers and also detected by DHDP. However, some are not (\figref{outliers} (d) (k) (l) (o)). \figref{outliers} (d) is a good example.  Spatially its probability is high because it is on one major activity shown in \figref{forumPatterns} P7. However, if its temporal information is considered, our method gives a low probability because its timing is very different from the observations in \figref{forumPatterns} P7. In contrast, DHDP gives a high probability because it first identifies the segment in which this trajectory is, then computes the probability based on the activities computed within the segment and the segments before. Since \figref{forumPatterns} P7 and \figref{outliers} (d) are in the same segment, a high probability is given. The result is caused by the fact that DHDP models progressions between segments but the temporal information within a segment is not modeled. Meanwhile, MOTIF reports a higher probability on \figref{outliers} (d). However, it suffers from the situation explained by the low likelihoods in \tabref{pwll} Left. When a continuous chunk of data are missing, there is a void spanning a short period in the training data, which causes low probabilities on any observations in the time span. This kind of temporal information loss leads to false alarms for anomaly detections (all our testing data report low probabilities). In our method, if an activity is seen before and after the void, it will be inferred that there is a probability that the activity also exists in the middle by putting a Gaussian over it. Even if the activity only appears before or after the missing data, the Gaussian there prevents the probability from decreasing as quickly as it does in MOTIF.
\section{Limitation and Conclusions}

For performance comparison, we tried to run three models and stopped them once satisfactory activities were computed and compared the time. Our method is approximately the same as DHDP and can be slightly slower than MOTIF depending on the dataset. But we did not use larger datasets because although being able to report good likelihoods, sampling is in general slow for largest datasets and it applies to all three models. So we focused on experiments that show the differences between our model and the other two. Also, we find the data is abundant in terms of activities where a random sampling suffices to reveal all activities. Quicker methods for training such as variational inference \cite{wang_path_2016} or parallel sampling can be employed in future.

In summary, we propose a new non-parametric hierarchical Bayesian model with a new hybrid sampling strategy for the posterior estimation for activity analysis. Its unique feature in time modeling provides better likelihoods, correctness/completeness and anomaly detection, which makes it a good alternative to existing models. We have shown its effectiveness on multiple datasets.

\bibliographystyle{splncs}
\bibliography{reference}

\begin{thebibliography}{10}

\bibitem{zhou_crowd_2010}
Zhou, S., Chen, D., Cai, W., Luo, L., Low, M.Y.H., Tian, F., Tay, V.S.H., Ong,
  D.W.S., Hamilton, B.D.:
\newblock Crowd {Modeling} and {Simulation} {Technologies}.
\newblock ACM Trans. Model. Comput. Simul. \textbf{20}(4) (November 2010)
  20:1--20:35

\bibitem{ali_floor_2008}
Ali, S., Shah, M.:
\newblock Floor {Fields} for {Tracking} in {High} {Density} {Crowd} {Scenes}.
\newblock In Forsyth, D., Torr, P., Zisserman, A., eds.: Computer {Vision} –
  {ECCV} 2008. Number 5303 in Lecture {Notes} in {Computer} {Science}.
\newblock Springer Berlin Heidelberg (October 2008)  1--14 DOI:
  10.1007/978-3-540-88688-4\_1.

\bibitem{antonini_behavioral_2006}
Antonini, G., Martinez, S.V., Bierlaire, M., Thiran, J.P.:
\newblock Behavioral {Priors} for {Detection} and {Tracking} of {Pedestrians}
  in {Video} {Sequences}.
\newblock Int J Comput Vision \textbf{69}(2) (May 2006)  159--180

\bibitem{emonet_extracting_2011}
Emonet, R., Varadarajan, J., Odobez, J.:
\newblock Extracting and locating temporal motifs in video scenes using a
  hierarchical non parametric {Bayesian} model.
\newblock In: 2011 {IEEE} {Conference} on {Computer} {Vision} and {Pattern}
  {Recognition} ({CVPR}). (June 2011)  3233--3240

\bibitem{zhou_learning_2014}
Zhou, B., Tang, X., Wang, X.:
\newblock Learning {Collective} {Crowd} {Behaviors} with {Dynamic}
  {Pedestrian}-{Agents}.
\newblock Int J Comput Vis \textbf{111}(1) (June 2014)  50--68

\bibitem{wang_trajectory_2008}
Wang, X., Ma, K.T., Ng, G.W., Grimson, W.:
\newblock Trajectory analysis and semantic region modeling using a
  nonparametric {Bayesian} model.
\newblock In: {IEEE} {Conference} on {Computer} {Vision} and {Pattern}
  {Recognition}, 2008. {CVPR} 2008. (June 2008)  1--8

\bibitem{wang_unsupervised_2009}
Wang, X., Ma, X., Grimson, W.:
\newblock Unsupervised {Activity} {Perception} in {Crowded} and {Complicated}
  {Scenes} {Using} {Hierarchical} {Bayesian} {Models}.
\newblock IEEE Trans. Patt. Anal. Machine Intel. \textbf{31}(3) (2009)
  539--555

\bibitem{stauffer_learning_2000}
Stauffer, C., Grimson, W.E.L.:
\newblock Learning {Patterns} of {Activity} {Using} {Real}-{Time} {Tracking}.
\newblock IEEE TRANSACTIONS ON PATTERN ANALYSIS AND MACHINE IN℡LIGENCE
  \textbf{22} (2000)  747--757

\bibitem{oliver_bayesian_2000}
Oliver, N., Rosario, B., Pentland, A.:
\newblock A {Bayesian} computer vision system for modeling human interactions.
\newblock IEEE Transactions on Pattern Analysis and Machine Intelligence
  \textbf{22}(8) (August 2000)  831--843

\bibitem{zelnik-manor_event-based_2001}
Zelnik-Manor, L., Irani, M.:
\newblock Event-based analysis of video.
\newblock In: Proceedings of the 2001 {IEEE} {Computer} {Society} {Conference}
  on {Computer} {Vision} and {Pattern} {Recognition}, 2001. {CVPR} 2001.
  Volume~2. (2001)  II--123--II--130 vol.2

\bibitem{zhong_detecting_2004}
Zhong, H., Shi, J., Visontai, M.:
\newblock Detecting unusual activity in video.
\newblock In: Proceedings of the 2004 {IEEE} {Computer} {Society} {Conference}
  on {Computer} {Vision} and {Pattern} {Recognition}, 2004. {CVPR} 2004.
  Volume~2. (June 2004)  II--819--II--826 Vol.2

\bibitem{lin_learning_2009}
Lin, D., Grimson, E., Fisher, J.:
\newblock Learning visual flows: {A} {Lie} algebraic approach.
\newblock In: {IEEE} {Conference} on {Computer} {Vision} and {Pattern}
  {Recognition}, 2009. {CVPR} 2009. (June 2009)  747--754

\bibitem{yi_understanding_2015}
Yi, S., Li, H., Wang, X.:
\newblock Understanding pedestrian behaviors from stationary crowd groups.
\newblock In: 2015 {IEEE} {Conference} on {Computer} {Vision} and {Pattern}
  {Recognition} ({CVPR}). (June 2015)  3488--3496

\bibitem{kitani_activity_2012}
Kitani, K.M., Ziebart, B.D., Bagnell, J.A., Hebert, M.:
\newblock Activity {Forecasting}.
\newblock In Fitzgibbon, A., Lazebnik, S., Perona, P., Sato, Y., Schmid, C.,
  eds.: Computer {Vision} – {ECCV} 2012. Number 7575 in Lecture {Notes} in
  {Computer} {Science}.
\newblock Springer Berlin Heidelberg (October 2012)  201--214 DOI:
  10.1007/978-3-642-33765-9\_15.

\bibitem{xie_inferring_2013}
Xie, D., Todorovic, S., Zhu, S.C.:
\newblock Inferring "{Dark} {Matter}" and "{Dark} {Energy}" from {Videos}.
\newblock In: 2013 {IEEE} {International} {Conference} on {Computer} {Vision}
  ({ICCV}). (December 2013)  2224--2231

\bibitem{wang_trajectory_2011}
Wang, X., Ma, K.T., Ng, G.W., Grimson, W.E.L.:
\newblock Trajectory {Analysis} and {Semantic} {Region} {Modeling} {Using}
  {Nonparametric} {Hierarchical} {Bayesian} {Models}.
\newblock Int J Comput Vis \textbf{95}(3) (May 2011)  287--312

\bibitem{varadarajan_sequential_2012}
Varadarajan, J., Emonet, R., Odobez, J.M.:
\newblock A {Sequential} {Topic} {Model} for {Mining} {Recurrent} {Activities}
  from {Long} {Term} {Video} {Logs}.
\newblock Int J Comput Vis \textbf{103}(1) (December 2012)  100--126

\bibitem{fei-fei_bayesian_2005}
Fei-Fei, L., Perona, P.:
\newblock A {Bayesian} hierarchical model for learning natural scene
  categories.
\newblock In: {IEEE} {Computer} {Society} {Conference} on {Computer} {Vision}
  and {Pattern} {Recognition}, 2005. {CVPR} 2005. Volume~2. (June 2005)
  524--531 vol. 2

\bibitem{sudderth_describing_2007}
Sudderth, E.B., Torralba, A., Freeman, W.T., Willsky, A.S.:
\newblock Describing {Visual} {Scenes} {Using} {Transformed} {Objects} and
  {Parts}.
\newblock Int J Comput Vis \textbf{77}(1-3) (2007)  291--330

\bibitem{sivic_discovering_2005}
Sivic, J., Russell, B.C., Efros, A.A., Zisserman, A., Freeman, W.T.:
\newblock Discovering object categories in image collections.
\newblock ICCV 2005 (2005)

\bibitem{niebles_unsupervised_2008}
Niebles, J.C., Wang, H., Fei-Fei, L.:
\newblock Unsupervised {Learning} of {Human} {Action} {Categories} {Using}
  {Spatial}-{Temporal} {Words}.
\newblock Int. J. Comp. Vision \textbf{79}(3) (2008)  299--318

\bibitem{kaufman_finding_2005}
Kaufman, L., Rousseeuw, P.J.:
\newblock Finding {Groups} in {Data}: {An} {Introduction} to {Cluster}
  {Analysis}.
\newblock Wiley-Interscience (2005)

\bibitem{emonet_temporal_2014}
Emonet, R., Varadarajan, J., Odobez, J.M.:
\newblock Temporal {Analysis} of {Motif} {Mixtures} {Using} {Dirichlet}
  {Processes}.
\newblock IEEE Transactions on Pattern Analysis and Machine Intelligence
  \textbf{36}(1) (January 2014)  140--156

\bibitem{teh_hierarchical_2006}
Teh, Y.W., Jordan, M.I., Beal, M.J., Blei, D.M.:
\newblock Hierarchical {Dirichlet} {Processes}.
\newblock J. Am. Stat. Assoc. \textbf{101}(476) (2006)  1566--1581

\bibitem{dubey_non-parametric_2012}
Dubey, A., Hefny, A., Williamson, S., Xing, E.P.:
\newblock A non-parametric mixture model for topic modeling over time.
\newblock arXiv:1208.4411 [stat] (August 2012) arXiv: 1208.4411.

\bibitem{wang_split-merge_2012}
Wang, C., Blei, D.M.:
\newblock A {Split}-{Merge} {MCMC} {Algorithm} for the {Hierarchical}
  {Dirichlet} {Process}.
\newblock arXiv:1201.1657 [cs, stat] (January 2012) arXiv: 1201.1657.

\bibitem{lin_construction_2010}
Lin, D., Grimson, E., Fisher, J.W.:
\newblock Construction of {Dependent} {Dirichlet} {Processes} based on
  {Poisson} {Processes}.
\newblock In Lafferty, J.D., Williams, C.K.I., Shawe-Taylor, J., Zemel, R.S.,
  Culotta, A., eds.: Advances in {Neural} {Information} {Processing} {Systems}
  23.
\newblock Curran Associates, Inc. (2010)  1396--1404

\bibitem{blei_distance_2011}
Blei, D.M., Frazier, P.I.:
\newblock Distance {Dependent} {Chinese} {Restaurant} {Processes}.
\newblock J. Mach. Learn. Res. \textbf{12} (November 2011)  2461--2488

\bibitem{wang_topics_2006}
Wang, X., McCallum, A.:
\newblock Topics over {Time}: {A} non-{Markov} {Continuous}-time {Model} of
  {Topical} {Trends}.
\newblock In: Proceedings of the 12th {ACM} {SIGKDD} {International}
  {Conference} on {Knowledge} {Discovery} and {Data} {Mining}. {KDD} '06, New
  York, NY, USA, ACM (2006)  424--433

\bibitem{sethuraman_constructive_1994}
Sethuraman, J.:
\newblock A constructive definition of {Dirichlet} priors.
\newblock Statistica Sinica \textbf{4} (1994)  639--650

\bibitem{hoffman_stochastic_2013}
Hoffman, M.D., Blei, D.M., Wang, C., Paisley, J.:
\newblock Stochastic variational inference.
\newblock Journal of Machine Learning Research \textbf{14}(1) (2013)
  1303--1347

\bibitem{chang_parallel_2014}
Chang, J., Fisher, J.W.:
\newblock Parallel {Sampling} of {HDPs} using {Sub}-{Cluster} {Splits}.
\newblock (2014)

\bibitem{hughes_effective_2012}
Hughes, M.C., Fox, E.B., Sudderth, E.B.:
\newblock Effective {Split}-{Merge} {Monte} {Carlo} {Methods} for
  {Nonparametric} {Models} of {Sequential} {Data}.
\newblock (2012)

\bibitem{jain_split-merge_2000}
Jain, S., Neal, R.:
\newblock A {Split}-{Merge} {Markov} {Chain} {Monte} {Carlo} {Procedure} for
  the {Dirichlet} {Process} {Mixture} {Model}.
\newblock Journal of Computational and Graphical Statistics \textbf{13} (2000)
  158--182

\bibitem{dahl_sequentially-allocated_2005}
Dahl, D.B.:
\newblock Sequentially-{Allocated} {Merge}-{Split} {Sampler} for {Conjugate}
  and {Nonconjugate} {Dirichlet} {Process} {Mixture} {Models}.
\newblock (2005)

\bibitem{majecka_statistical_2009}
Majecka, B.:
\newblock Statistical models of pedestrian behaviour in the {Forum}.
\newblock {MSc} {Dissertation}, School of Informatics, University of Edinburgh,
  Edinburgh (2009)

\bibitem{luber_socially-aware_2012}
Luber, M., Spinello, L., Silva, J., Arras, K.O.:
\newblock Socially-aware robot navigation: {A} learning approach.
\newblock In: 2012 {IEEE}/{RSJ} {International} {Conference} on {Intelligent}
  {Robots} and {Systems} ({IROS}). (October 2012)  902--907

\bibitem{almingol_learning_2013}
Almingol, J., Montesano, L., Lopes, M.:
\newblock Learning {Multiple} {Behaviors} from {Unlabeled} {Demonstrations} in
  a {Latent} {Controller} {Space}.
\newblock (2013)  136--144

\bibitem{wang_path_2016}
Wang, H., Ondřej, J., O'Sullivan, C.:
\newblock Path {Patterns}: {Analyzing} and {Comparing} {Real} and {Simulated}
  {Crowds}.
\newblock In: Proceedings of the 20th {ACM} {SIGGRAPH} {Symposium} on
  {Interactive} 3D {Graphics} and {Games}. I3D '16, New York, NY, USA, ACM
  (2016)  49--57

\end{thebibliography}


\begin{thebibliography}{1}

\bibitem{teh_hierarchical_2006}
Teh, Y.W., Jordan, M.I., Beal, M.J., Blei, D.M.:
\newblock Hierarchical {Dirichlet} {Processes}.
\newblock J. Am. Stat. Assoc. \textbf{101}(476) (2006)  1566--1581

\end{thebibliography}

\end{document}


\pagestyle{headings}
\mainmatter

\title{Supplementary Material of Globally Continuous and Non-Markovian Crowd Activity Analysis from Videos}

\titlerunning{Globally Continuous and Non-Markovian Crowd  Activity Analysis from Videos}

\authorrunning{He Wang and Carol O'Sullivan}

\author{He Wang\inst{1,2} and Carol O'Sullivan\inst{1,3}\thanks{Correspondence Author}}
\institute{Disney Research Los Angeles\thanks{This work is mostly done by the authors when they were with Disney Research Los Angeles.}, United States of America
\and
University of Leeds, United Kingdoms \\
\email{realcrane@gmail.com}
\and 
Trinity College Dublin, Ireland \\
\email{carol.osullivan@scss.tcd.ie}}

\maketitle

\section{Generative Process of STHDP}
\begin{figure}
\centering
  \begin{minipage}[b]{0.4\textwidth}
    \includegraphics[width=\textwidth]{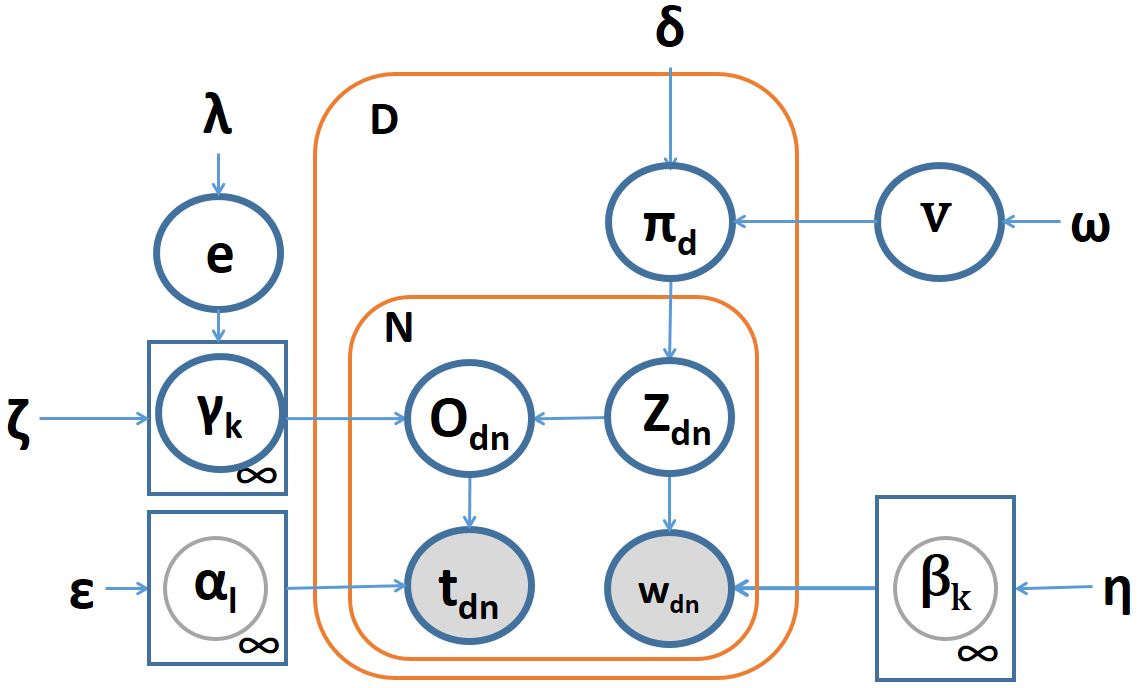}
    \caption{STHDP model}
    \label{fig:STHDP}
  \end{minipage}
  \begin{minipage}[b]{0.5\textwidth}
    \includegraphics[width=\textwidth]{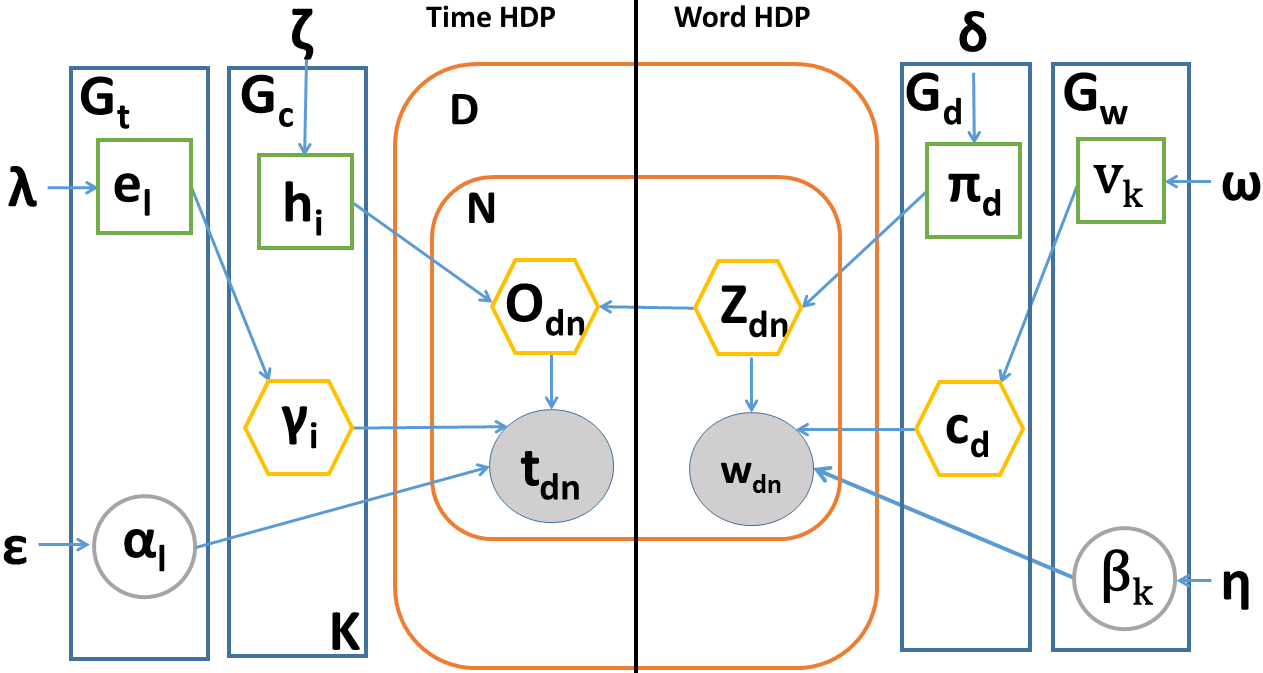}
    \caption{Model used for sampling.}
    \label{fig:STHDPSampling}
  \end{minipage}
\end{figure}

We first again show the STHDP model in \figref{STHDP}.

The generative process of \figref{STHDP} is explained as follows:

\begin{enumerate}
\item Sample a corpus-level time base distribution, $e | \lambda \sim GEM(\lambda)$
\item Sample a corpus-level word base distribution, $v | \omega \sim GEM(\omega)$
\item For each corpus-level word topic $k$:
\begin{enumerate}
	\item Sample a distribution over words, $\beta_k | \eta \sim Dirichlet(\eta)$
	\item Sample a word-topic-specific distribution over time topics, $\gamma_k | e, \zeta \sim DP(\zeta, e)$
\end{enumerate}
\item For each time topic $l$:
	\begin{enumerate}
	\item Sample a distribution over time, $\alpha_l | \Gamma \sim $ Normal-Inverse-Gamma($\Gamma$)
	\end{enumerate}
\item For each document $d$:
	\begin{enumerate}
	\item Sample a distribution over topics, $\pi_d | v, \sigma \sim DP(\sigma, v)$
	\item For each word $n$:
		\begin{enumerate}
		\item Sample a word topic indicator, $z_{d_n} | \pi_d \sim \pi_d$
		\item Sample a word $w_{dn} | \beta_{z_{d_n}} \sim Mult(\beta_{z_{d_n}})$
		\item Sample a time topic indicator, $o_{d_n} | z_{d_n}, \gamma \sim \gamma_{z_{d_n}}$
		\item Sample a time word $t_{dn} | \alpha_{o_{d_n}} \sim Normal(\alpha_{o_{d_n}})$
		\end{enumerate}
	\end{enumerate}
\end{enumerate}
\section{Gibbs Sampling for STHDP}
Based on \figref{STHDPSampling}, we only explain the modified Chinese Restaurant Franchise scheme here, as we fix the word HDP while running Chinese Restaurant Franchise (CRF) on the time HDP as in \cite{teh_hierarchical_2006}. Following the naming convention in CRF, word topics and time topics are called \textbf{\textit{word\ dishes}} and \textbf{\textit{time\ dishes}}. Word documents are called \textbf{\textit{restaurants}} and the set of time stamps associated with one word topic is called a \textbf{\textit{time\ restaurant}}. A list of auxiliary variables are given in \tabref{crfVar}. Also, we use superscript to exclude data samples. For instance, $z_{j}^{-ji}$ means the set of all table indicators in restaurant j excluding $w_{ji}$. Bold fonts means the whole set of some quartile, for instance, $\textbf{l}$ means all time dish indices. We also use dots as summation. $m_{\cdot k}$ is the number of word tables serving dish k.

\begin{table}
\caption{Variables in CRF}
\label{tab:crfVar}
\begin{tabular}{|p{0.5cm}|p{7cm}|}
\hline
$v_w$ & a word in the vocabulary \\
\hline
$V_w$ & the size of the vocabulary \\
\hline
$w_{ji}$ & the $i$th word in restaurant j \\
\hline
$t_{ji}$ & the $i$th time word in restaurant j \\
\hline
$n_{ji}$ & the number of words in restaurant j at table i\\
\hline
$n_{j\cdot}$ & the number of words in restaurant j\\
\hline
$z_{ji}$ & the table indicator of the $i$th word in restaurant j\\
\hline
$k_{ji}$ & the dish indicator of the $i$th word table in restaurant j\\
\hline
$m_{jk}$ & the number of word tables in restaurant j serving dish k\\
\hline
$m_{j\cdot}$ & the number of word tables in restaurant j\\
\hline
$K$ & the number of word dishes\\
\hline
$s_{ji}$ & the number of time words in time restaurant j at time table i\\
\hline
$s_{j\cdot}$ & the number of time words in time restaurant j\\
\hline
$d_{jl}$ & the number of time tables in time restaurant j serving time dish l\\
\hline
$d_{j\cdot}$ & the number of tables  in time restaurant j\\
\hline
$o_{ji}$ & the table indicator of the $i$th time word in time restaurant j\\
\hline
$l_{ji}$ & the time dish indicator of the $i$th table in time restaurant j\\
\hline
\end{tabular}
\end{table}

\subsubsection{Sampling Word Tables}
The full conditional of a table indicator, $z_{ji}$, for a word, $w_{ji}$, given all other words is:
\begin{equation}
\label{eq:wordTable}
\begin{split}
&p(z_{ji} = z, w_{ji}, t_{ji} | \textbf{z}^{-ji}, \textbf{w}^{-ji}, \textbf{t}^{-ji}, \textbf{k}, \textbf{o}^{-ji}, \textbf{l})  = \\
&p(z_{ji} = z | \textbf{z}^{-ji}) \\
&p(w_{ji} | t_{ji}, z_{ji} = z, k_{jz} = k, \textbf{w}^{-ji}, \textbf{z}^{-ji}, \textbf{k}, \textbf{t}^{-ji}, \textbf{o}^{-ji}, \textbf{l}) \\
&p(t_{ji} | z_{ji}= z, k_{jz} = k, \textbf{t}^{-ji}, \textbf{o}^{-ji}, \textbf{l})
\end{split}
\end{equation}
where
\begin{equation}
p(z_{ji} = z | \textbf{z}^{-ji}) \propto 
\begin{cases}
n_{jz} & \quad \text{if z is an existing table} \\
\delta & \quad \text{otherwise}
\end{cases}
\end{equation}
\begin{equation}
\begin{split}
p(w_{ji} | t_{ji}, z_{ji} = z, &k_{jz} = k, \textbf{w}^{-ji}, \textbf{z}^{-ji}, \textbf{k}, \textbf{t}^{-ji}, \textbf{o}^{-ji}, \textbf{l}) \\
&\propto 
\begin{cases}
f_{k_z}(w_{ji})  & \quad \text{if z exists} \\
m_{\cdot k} \ f_{k}(w_{ji}) & \quad \text{else if k exists} \\
\omega \ f_{k_{new}}(w_{ji}) & \quad \text{otherwise}
\end{cases}
\end{split}
\end{equation}
where $f$ is the conditional density of $w_{ji}$ given all other variables. 
$p(t_{ji} | z_{ji}= z, k_{jz} = k, \textbf{t}^{-ji}, \textbf{o}^{-ji}, \bf{l})$ is the extra term from the time HDP that needs special treatment.  If, for every word, we do sampling conditioned on its time word in the time HDP , it is very slow. So we marginalize over all time tables in the time restaurant.
\begin{equation}
\label{eq:timeTopic}
\begin{split}
p(t_{ji} | &z_{ji} = z, k_{jz} = k, \textbf{t}^{-ji}, \textbf{o}^{-ji}, \textbf{l}) = \\
&\sum_{o = 1}^{d_{j\cdot}} p(o_{ji} = o | z_{ji}= z, k_{jz} = k,  \textbf{t}^{-ji}, \textbf{o}^{-ji}) \\
&p(t_{ji} | o_{ji} = o, l_{jo} = l, \textbf{l})
\end{split}
\end{equation}
\begin{equation}
\begin{split}
 p(o_{ji} = o |  z_{ji}= z, k_{jz} = k, &\textbf{t}^{-ji}, \textbf{o}^{-ji}) \\
&\propto
\begin{cases}
s_{ji} & \quad \text{if o exists} \\
\zeta & \quad \text{otherwise}
\end{cases}
\end{split}
\end{equation}
\begin{equation}
p(t_{ji} | o_{ji} = o, l_{jo_{ji}} = l, \textbf{l}) \propto
\begin{cases}
g_{l_o}(t_{ji}) & \quad \text{if o exists} \\
d_{\cdot l} \ g_{l}(t_{ji}) & \quad \text{else if} \ l \ \text{exists} \\
\varepsilon \ g_{l_{new}}(t_{ji}) & \quad \text{otherwise}
\end{cases}
\end{equation}
where $g$ is the posterior predictive distribution of a Gaussian, a t-Distribution. 

\subsubsection{Sampling Word Dishes}
Sampling a word topic for a word table $z$ in restaurant $j$, with the associated words $\bf{w}_{jz}$ and  time words $\bf{t}_{jz}$, follows:

\begin{equation}
\begin{split}
p(k_{jz} &= k, \textbf{w}_{jz}, \textbf{t}_{jz} | \textbf{w}^{-jz}, \\
&\textbf{t}^{-jz}, \textbf{z}^{-jz}, \textbf{k}^{-jz}, \textbf{o}^{-jz}, \textbf{l}^{-jz}) \\
&\propto 
\begin{cases}
m_{\cdot k}^{-jz} p(\textbf{w}_{jz} | \bullet ) p(\textbf{t}_{jz} | \bullet ) & \ \text{if k exists} \\
\omega \  p(\textbf{w}_{jz} | \bullet) p(\textbf{t}_{jz} | \bullet) & \ \text{otherwise}
\end{cases}
\end{split}
\end{equation}
where $\bullet$ means all the other variables the distribution is conditioned on. $p(\textbf{w}_{jz} | \bullet ) = f_k(\textbf{w}_{jz})$. To fully compute $p(\textbf{t}_{jz} | \bullet) = p(\textbf{t}_{jz} | k_z = k, \textbf{o}^{-jz}, \textbf{l}^{-jz})$ is too expensive because table $z$ might have many words. So we randomly sample a number of them to compute $\hat{p}(\textbf{t}_{jz} | \bullet)$ as an approximation, which can be computed by \eqref{timeTopic}.

\section{Additional Results}
We show some additional patterns in the Forum dataset and TrainStation dataset in \figref{forumAdditional} and \figref{trainStationAdditional}.
\InsertFig{\includegraphics[scale=0.6]{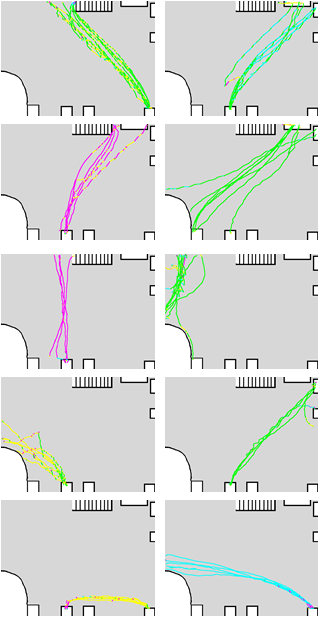}}{Additional Patterns in Forum dataset}{forumAdditional}{th}

\InsertFig{\includegraphics[scale=0.6]{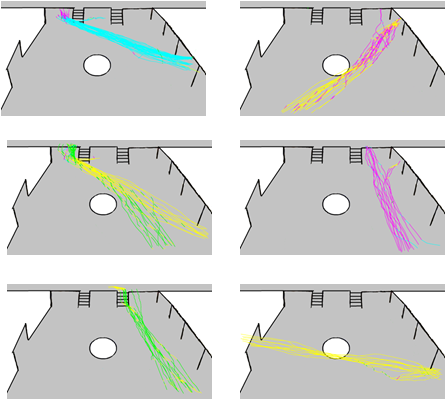}}{Additional Patterns in TrainStationAdditional dataset}{trainStationAdditional}{th}

{\small
\bibliographystyle{splncs}
\bibliography{reference}
}